\documentclass[letterpaper]{article} 
\usepackage{aaai25}  
\usepackage{times}  
\usepackage{helvet}  
\usepackage{courier}  
\usepackage[hyphens]{url}  
\usepackage{graphicx} 
\usepackage{natbib}  
\usepackage{caption} 
\usepackage{algorithm}
\usepackage{algorithmic}
\usepackage{helvet}  
\usepackage{courier}  
\usepackage[hyphens]{url}  
\usepackage{graphicx} 
\usepackage{natbib}  
\usepackage{caption} 
\usepackage{amsmath}
\usepackage{booktabs}
\usepackage{xcolor}
\usepackage{bm}
\usepackage{amsmath}
\usepackage{booktabs}
\usepackage{xcolor}
\usepackage{colortbl}
\usepackage{pifont}
\usepackage{xcolor}
\usepackage{colortbl}
\usepackage{array}
\usepackage{makecell}
\usepackage{titletoc}
\usepackage{multirow}
\usepackage{graphicx} 
\usepackage{colortbl}
\newcommand{\cmark}{\ding{51}}%
\newcommand{\xmark}{\ding{55}}%
\usepackage{graphicx} 

\usepackage{xr}

\usepackage{xspace}
\makeatletter
\DeclareRobustCommand\onedot{\futurelet\@let@token\@onedot}
\def\@onedot{\ifx\@let@token.\else.\null\fi\xspace}

\def\ie{\emph{i.e}\onedot}

\makeatother

\usepackage{algorithm}
\usepackage{algorithmic}
\usepackage{amsfonts} 
\usepackage{newfloat}
\usepackage{listings}
\usepackage{newfloat}
\usepackage{listings}
\DeclareCaptionStyle{ruled}{labelfont=normalfont,labelsep=colon,strut=off} 
\floatstyle{ruled}
\newfloat{listing}{tb}{lst}{}
\floatname{listing}{Listing}
\definecolor{ourscolor}{HTML}{c2d1e5}

\title{Ultra-High Resolution Segmentation via Boundary-Enhanced Patch-Merging Transformer}
\author {
    Haopeng Sun\textsuperscript{\rm 1,\rm 2} \thanks{Equal contribution.},
    Yingwei Zhang\textsuperscript{\rm 1,\rm 2} \footnotemark[1],
    Lumin Xu\textsuperscript{\rm 4},
    Sheng Jin\textsuperscript{\rm 5,\rm 6},
    Yiqiang Chen\textsuperscript{\rm 1,\rm 2,\rm 3} \thanks{Corresponding author.}
}

\affiliations {
    \textsuperscript{\rm 1}Institute of Computing Technology, Chinese Academy of Sciences \\
    \textsuperscript{\rm 2}University of Chinese Academy of Sciences\\
    \textsuperscript{\rm 3}Peng Cheng Laboratory    \quad \quad
    \textsuperscript{\rm 4}The Chinese University of Hong Kong \\
    \textsuperscript{\rm 5}The University of Hong Kong    \quad \quad
    \textsuperscript{\rm 6}SenseTime Research and Tetras.AI \\
    sunhaopeng22s@ict.ac.cn, zhangyingwei@ict.ac.cn, luminxu@link.cuhk.edu.hk, jinsheng@tetras.ai, yqchen@ict.ac.cn
}

\usepackage{bibentry}

\begin{document}

\maketitle

\begin{abstract}
Segmentation of ultra-high resolution (UHR) images is a critical task with numerous applications, yet it poses significant challenges due to high spatial resolution and rich fine details. Recent approaches adopt a dual-branch architecture, where a global branch learns long-range contextual information and a local branch captures fine details. However, they struggle to handle the conflict between global and local information while adding significant extra computational cost. Inspired by the human visual system's ability to rapidly orient attention to important areas with fine details and filter out irrelevant information, we propose a novel UHR segmentation method called Boundary-enhanced Patch-merging Transformer (BPT). BPT consists of two key components: (1) Patch-Merging Transformer (PMT) for dynamically allocating tokens to informative regions to acquire global and local representations, and (2) Boundary-Enhanced Module (BEM) that leverages boundary information to enrich fine details. Extensive experiments on multiple UHR image segmentation benchmarks demonstrate that our BPT outperforms previous state-of-the-art methods without introducing extra computational overhead. Codes will be released to facilitate research.
\end{abstract}

%


\section{Introduction}
With the advancement of remote sensing technology, the acquisition of abundant ultra-high resolution (UHR) geospatial images has become possible~\cite{liu2023seeing,CMSCGC,SSGCC,li2024sglp,li2024comae}. Semantic segmentation of UHR geospatial images has opened new horizons in computer vision, playing an increasingly important role in earth sciences and urban applications such as disaster control, environmental monitoring, land resource management, conservation, and urban planning~\cite{chen2019collaborative,ji2023guided,ji2023ultra,zhu2024exact}. 
The main challenge of this task is how to balance the semantic dispersion of high-resolution targets in the small receptive field and the loss of high-precision details in the large receptive field~\cite{zhao2018icnet}, as well as the dilemma of trading off computation cost and segmentation accuracy.


\begin{figure*}[t]
    \centering
    \includegraphics[width=0.90\linewidth]{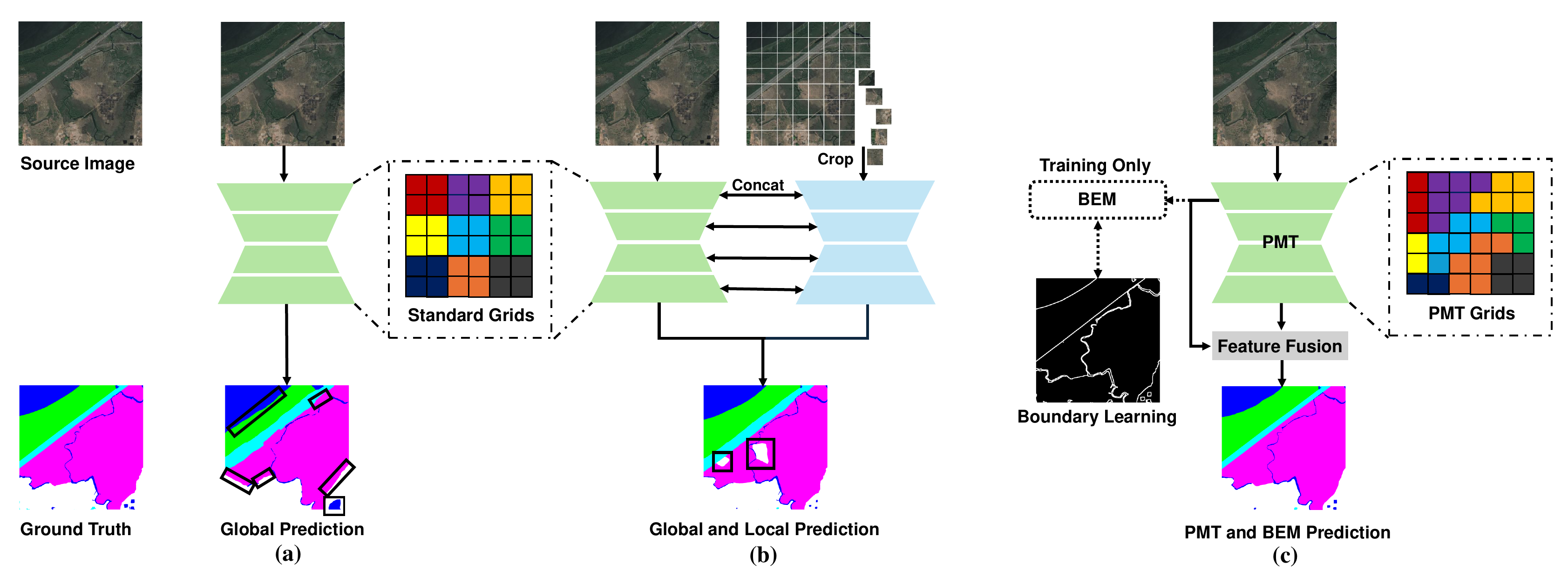}
    \caption{(a) Existing methods representing images as standard grids of pixels are sub-optimal for UHR segmentation. (b) Dual-branch framework preserves both global and local information at the cost of increased computation. (c) Our proposed model captures both global and local information by dynamically allocating tokens to informative regions (PMT) and leveraging boundary information (BEM).}
    \label{fig:teaser}  
    \vspace{-0.4cm}
\end{figure*}

Convolutional neural networks~\cite{ronneberger2015u,zhao2017pyramid,chen2018encoder,yu2018bisenet,jiang2023latent,jiang2024pgiun,zhang2024pseudo,sun2024program} and vision transformers~\cite{cheng2024sptsequenceprompttransformer,ji2023ultra,shen2023pbsl,shen2023triplet,lu2023towards,lu2024scaling,Lu2024DrivingRecon,shen2023git,long2024generative,hu2024monobox,xie2024automated,xie2024knowledge,xie2024video,zhou2023improving,zhou2024metagpt} have succeeded in representing images as uniform grids of pixels or patches of fixed size. However, as shown in Fig.~\ref{fig:teaser} (a) and (b), the previous works are sub-optimal for high-resolution remote sensing imagery analysis due to either overlooking fine-grained information~\cite{chen2019collaborative,cheng2020cascadepsp,zhang2024cf,zhang2023multi,tao2023dudb} or introducing substantial computational overhead~\cite{li2021contexts,ji2023guided,li2024cloud,li2024deviation,li2024translating}.
On the one hand, large areas of ocean contains global contextual information but little textual details can be represented by a single token for efficient global semantic learning. On the other hand, buildings and roads with fine-grained structures require more tokens to preserve details. 
Inspired by the human visual system which rapidly orients attention to important areas with fine details and filter out large amount of irrelevant information in complex large scenes, we propose a novel method termed \textbf{B}oundary-Enhanced \textbf{P}atch-Merging \textbf{T}ransformer (BPT) to tackle this problem. 
As shown in Fig.~\ref{fig:teaser} (c), 
the specifically designed Patch-Merging Transformer (PMT) adopts a dynamic patch merging approach, where vision tokens represent different regions with dynamic shapes and sizes.
It efficiently captures both global and local information by dynamically adopting a fine resolution for regions containing critical details while preserving global information. 


To further enhance performance, the recent approaches~\cite{guo2022isdnet,ji2023guided,ji2023ultra,liu2023seeing,he2024geolocation} for UHR images follow a dual-branch framework to preserve both global and local information. These methods involve two deep branches: one downsamples the entire image and extracts global information, while the other crops local patches feeding them to the network sequentially and merging their predictions to obtain local cues. 
However, the former branch loses local details leading to inaccurate edge segmentation, while the latter branch lacks global context resulting in semantic errors. 
Directly fusing features from different branches may lead to information confusion, and the introduction of two-branch structure increases the memory cost.

To tackle this problem, we propose the Boundary-Enhanced Module (BEM) to improve the edge of segmentation masks.
By introducing intermediate supervision on the low-level features, the boundary details are learned in high resolutions.
It is notable that the auxiliary boundary learning module can be discarded during inference, resulting in no additional computation overhead. 
In addition, to prevent the confusion caused by directly fusing global and local information, we propose a feature fusion module to adaptively weigh different information to obtain the more accurate features.

Extensive experiments on five public UHR image segmentation benchmark datasets demonstrate that BPT outperforms the previous state-of-the-art methods, validating the effectiveness of our proposed method.

Our main contributions can be summarized as follows:
\begin{itemize}
\item We propose a novel efficient UHR image segmentation method termed \textbf{B}oundary-Enhanced \textbf{P}atch-Merging \textbf{T}ransformer (BPT) to address the issue of global and local information fusion and strike the computation-accuracy balance. 

\item We propose the Patch-Merging Transformer (PMT) which dynamically represents remote sensing regions with tokens with various shape and size, capturing both the global contextual information and rich local details. 

\item We propose the Boundary-Enhanced Module (BEM) to integrate detailed boundary spatial information without introducing an extra time-consuming branch.

\item Extensive experiments on five public benchmark datasets demonstrate that the proposed method outperforms the existing approaches.
\end{itemize}

\begin{figure*}[ht]
    \centering
    \includegraphics[width=0.85\linewidth]{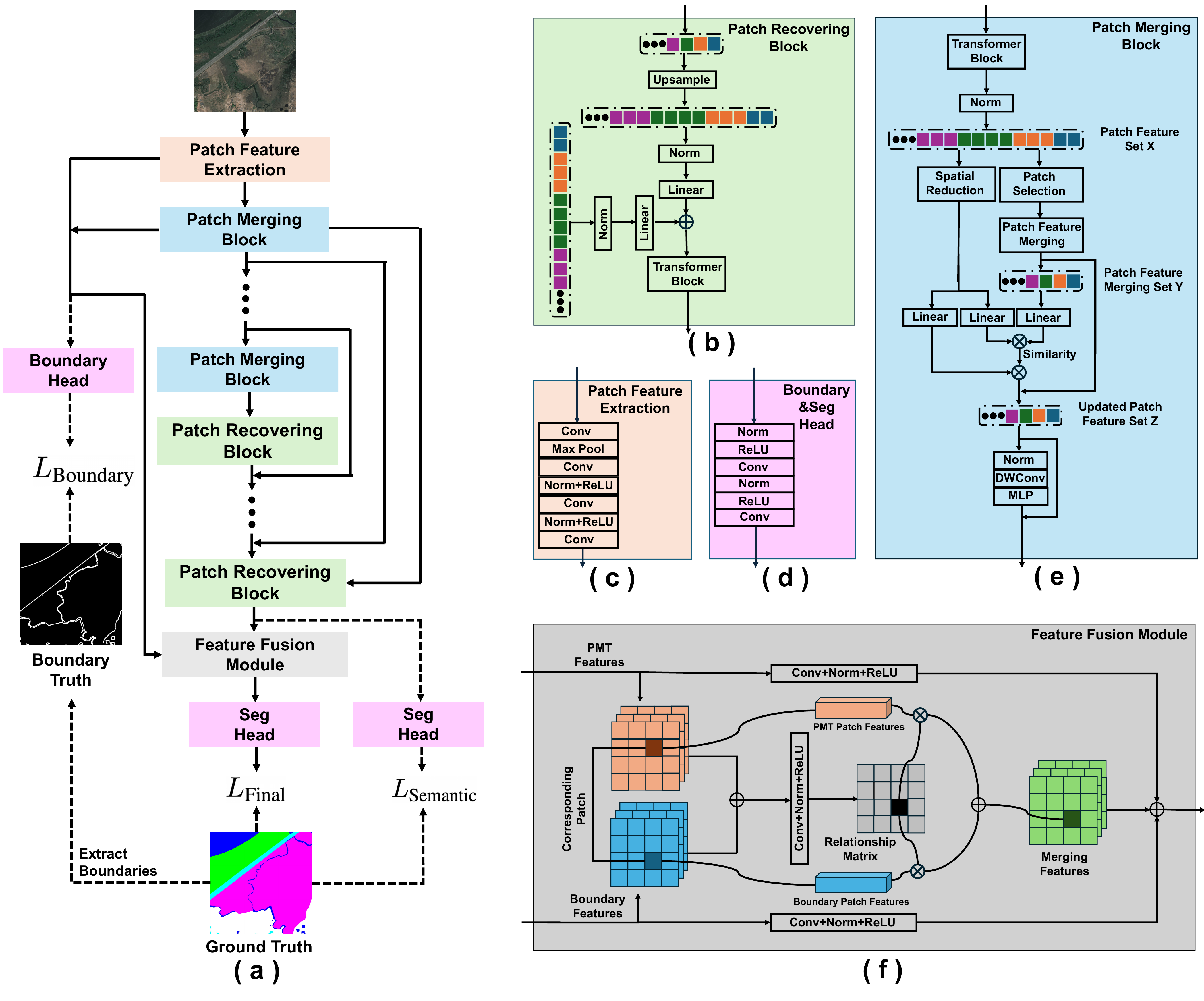}
    \caption{(a) Overview of \textbf{B}oundary-Enhanced \textbf{P}atch-Merging \textbf{T}ransformer (BPT), which consists of PMT and BEM. Dotted lines represent that only needed during the training phase. (b) Patch Recovering Block, (c) Patch Feature Extraction, (d) Boundary \& Seg Head, (e) Patch Merging Block, (f) Feature Fusion Module.
    }
    \label{fig:overview}   
    \vspace{-0.3cm}
\end{figure*}

\section{Related Work}

\subsection{Generic Image Segmentation}
Deep learning methods have significantly advanced the field of semantic segmentation, especially for natural images and daily photos. 
Early semantic segmentation models~\cite{Yuan2024,Yuan2024e,Yuan2024g} were primarily based on Fully Convolutional Networks (FCNs). FCN-based methods typically utilized an encoder to downsample images, thereby reducing spatial resolution while extracting high-level semantic features, and a decoder to upsample images, restoring spatial resolution while classifying each pixel. 
For instance, Deeplabv3~\cite{chen2018encoder} adopted an atrous spatial pyramid pooling module to capture long-range context, and PSPNet~\cite{zhao2017pyramid} devised a pyramid pooling strategy to capture both local and global context information. UNet~\cite{ronneberger2015u} proposed a symmetric encoder-decoder structure for efficient and precise segmentation, preserving detailed information and enhancing segmentation accuracy.
With the superior ability of Transformers to capture long-distance information, Transformer-based networks~\cite{xie2021segformer,cheng2021per,nie2024imputeformer,qianmaskfactory,yin2022camoformer,yin2023dformer,zhu2024misa,10587023,10723789} have become a new research focus in segmentation, with representative works including SegFormer~\cite{xie2021segformer} and MaskFormer~\cite{cheng2021per}. 
However, due to memory limitations, it is challenging to apply generic semantic segmentation methods to ultra-high resolution (UHR) images. Existing UHR image segmentation approaches generally downsample images to regular resolutions or crop images into small patches, processing them sequentially and merging their predictions. 
Cropping can lead to global semantic errors, while downsampling can result in inaccurate segmentation of details, thus producing suboptimal results. 
Additionally, existing networks commonly use uniform grids of pixels or fixed-size patches, which are suboptimal for high-resolution remote sensing imagery analysis. Instead, we propose
the Patch-Merging Transformer (PMT) to dynamically allocate vision tokens with varying shapes and sizes to represent different image regions, enhancing the segmentation performance while maintaining computational efficiency.

\subsection{Ultra-High Resolution Image Segmentation}
Advancements in photography and sensor technologies have increased the accessibility of Ultra-High Resolution (UHR) geospatial images, opening new horizons for the computer vision community. 
Most existing methods for UHR image segmentation utilize multi-branch networks to learn both global and local information. 
GLNet~\cite{chen2019collaborative} incorporated global and local information deeply in a two-stream branch manner. 
CascadePSP~\cite{cheng2020cascadepsp} proposed a multi-branch network to learn features at different scales, generating high-quality results. 
FCtL~\cite{li2021contexts} exploited three different cropping scales to fuse multi-scale feature information. 
ISDNet~\cite{guo2022isdnet} integrated shallow and deep networks to learn global and local information respectively. 
WSDNet~\cite{ji2023ultra} introduced a multi-level discrete wavelet transform into the global and local branches to reduce computational overhead. 
GPWFormer~\cite{ji2023guided} used a hybrid CNN-Transformer in a dual-branch style to efficiently harvest both low-level and high-level context. 
GeoAgent~\cite{liu2023seeing} employed a reinforcement learning network to dynamically adjust the size of patches for global context.
These existing methods often design complex multi-encoder-decoder streams and stages to gradually fuse global and local information, resulting in high memory requirements. 
In addition, directly merging these features can lead to information confusion and segmentation errors. 
To address these issues, we propose a novel efficient UHR image segmentation method called Boundary-Enhanced Patch-Merging Transformer (BPT). BPT adopts a single-branch structure capturing both global and local information by accounting for varying receptive field requirements in different instances. In addition, the auxiliary Boundary-Enhanced Module (BEM) leverages boundary information for learning fine details.

\section{Method}

\subsection{Overview}
We introduce our proposed method for ultra-high resolution (UHR) image segmentation, termed Boundary-Enhanced Patch-Merging Transformer (BPT). As shown in Fig.~\ref{fig:overview}, BPT comprises two main components: Patch-Merging Transformer (PMT) and Boundary-Enhanced Module (BEM). Each UHR image is first cropped into several patches. PMT dynamically merges patches based on instance sizes, extracting both global and local information for more accurate preliminary segmentation. Additionally, the BEM learns boundary information to guide the fusion of fine boundary details and local information, enhancing segmentation accuracy without extra time or memory consumption.

\subsection{Patch Merging Transformer (PMT)}

Our proposed Patch-Merging Transformer (PMT) is composed of Patch Feature Extraction (PFE), Patch Merging Block (PMB) and Patch Recovering Block (PRB). The input image is first uniformly divided in to small patches and processed through PFE to extract patch features. Then, the patches are automatically merged through PMB to reduce their number and thereby decrease memory consumption. Next, the merged patches are restored to the original number of patches through PRB and get the feature maps.

\subsubsection{Patch Feature Extraction (PFE).}
The input UHR image is evenly partitioned into image patches to reduce resolution. Unlike previous methods, we use smaller, more numerous patches for finer segmentation. We crop the source image into patches with a size of 32$\times$32 pixels. 
A higher number of patches is beneficial for preserving detailed information, and since our method can dynamically merge patches, it does not consume a large amount of memory.
In implementation, we adopt PVT block as the base transformer block, for its high computation efficiency.

\subsubsection{Patch Merging Block (PMB).} 

We employ the Patch Merging Block (PMB) to gradually fuse patches and extract deep semantic information.

\textbf{Patch Selection.} 
Inspired by~\cite{zeng2022not}, we apply the clustering algorithm to merge similar patch features, reducing the number of patches. Specifically, we use a variant of k-nearest neighbor based density peaks clustering algorithm (DPC-KNN). Given a set of patches $\bm{X}=[x_1, ..., x_i, ..., x_j, ...]$, we compute the local density $\rho$ of each patch based on its k-nearest neighbors:

\begin{equation}
\rho_i = \exp \left( -\frac{1}{k} \sum_{x_j \in \mathrm{KNN}(x_i)} \left\| x_i - x_j \right\|_2^2 \right), \label{eq:1}
\end{equation}
where $\mathrm{KNN}(x_i)$ denotes the k-nearest neighbors of patch $i$. $x_i$ and $x_j$ are their corresponding patch features. We then compute the distance indicator as the minimal distance between a patch and any other patch with higher local density. For the patch with the highest local density, its indicator is set as the maximal distance between it and any other patches:

\begin{equation}
\delta_i = \left\{
\begin{array}{ll}
\min_{j: \rho_j > \rho_i} \left\| x_i - x_j \right\|_2, & \text{if } \exists j \text{ s.t. } \rho_j > \rho_i \\
\max_j \left\| x_i - x_j \right\|_2, & \text{otherwise}
\end{array}
\right. \label{eq:2}
\end{equation}
where $\delta_i$ is the distance indicator and $\rho_i$ is the local density. We combine the local density and the distance indicator to score each patch as $\rho_i \times \delta_i$. Higher scores indicate higher potential as patch centers. We determine patch centers by selecting tokens with the highest scores and then assign other tokens to the nearest patch center based on feature distances.

\textbf{Patch Feature Merging.} 
To focus on the important image features when merging patch features, we utilize an importance-based patch feature merging strategy. Specifically, we use the importance score $P_j$ to represent the importance of each patch, estimated from the patch local density and distance indicator. The merged patch feature set $\bm{Y}=[y_1, y_2, ...]$ is calculated as:

\begin{equation}
y=\sum_{j \in C_i} Softmax(\rho_j \cdot \delta_j) x_j, \label{eq:3}
\end{equation}
where $C_i$ is the set of the $i$-th cluster, $x_j$ and $p_j$ are the original patch features and the corresponding importance score, respectively, and $y$ is the features of the merged patch.


\textbf{Patch Feature Updating.} We enhance the merged patch features by considering the relations between the patch features before and after merging. Formally, we compute the similarity matrix $S$ between the merged features (patch feature set after merging) and original features (patch feature set before merging) as:

\begin{equation}
Similarity(y_m,x_i)=S_{m,i} = \frac{\exp(W_q y_m \cdot W_k x_i )}{\sum_{n=1}^{N} \exp(W_q y_n \cdot W_k x_i)},
\end{equation}
where $x_i$ and $y_m$ are the $i$-th of $\bf{X}$ and $m$-th of $\bf{Y}$, $W_q$ and $W_k$ are weights of learned linear projections for patch features, and $N$ is the number of elements in $\bm{Y}$. 
The merged patch features are updated by adding a residual term capturing the details of the original features:
\begin{equation}
z_m = y_m + W_o \frac{\sum_{i=1}^{N_x} S_{m,i} W_v x_i}{\sum_{i=1}^{N_x} S_{m,i}},
\end{equation}
where $N_x$ is the number of $\bf{X}$, and $W_v$ and $W_o$ are learned weights to project merged features. 
In practice, we need to first transform tokens to feature maps before the projection process and perform the inverse transform after. We transform feature maps into patch features before merging and perform the inverse transformation afterward. Specifically, patch features correspond to specific positions in feature maps. The patches that undergo merging will have identical features, reducing the number of patches while expanding their regions. Patch features are inverse-transformed back into feature maps based on the position correspondence. Finally, the obtained feature maps are passed sequentially through Norm, DWConv, and MLP. 

\subsubsection{Patch Recovering Block (PRB).}
Patch Recovering Blocks (PRB) aggregate patches of different scales (ordinary and dynamically-merged patch features) and reconstruct output features for upsampling. PRB uses two linear layers and one PVT transformer block to reduce computation burdens. For each merged patch feature containing abstract semantics, PRB upsamples patches and recovers feature mapping based on its merging history. During token merging in PFM, we record positional correspondence between original and merged patch features. In PRB's upsampling process, these records copy merged patch features into corresponding upsampled patches, executed progressively until all patch features are aggregated.

\subsection{Boundary-Enhanced Module (BEM)}
We propose a Boundary-Enhanced Module (BEM) to guide low-level layers in learning boundary information~\cite{wu2022multi} and refine global information for more precise results. This process occurs only during training, avoiding additional memory or time consumption during inference.

\textbf{Boundary Prediction Task (BPT).} Downsampling and upsampling processes can lose the boundary information, degrading segmentation performance at boundaries. To preserve the fine boundary details, we introduce an auxiliary task guiding low-level layers to learn boundary prediction. 
The boundary prediction is formulated as a binary segmentation task. We use the Canny operator and Dilation operation to obtain the boundaries mask.

\textbf{Feature Fusion Module (FFM).} 
The PMT path is semantically accurate but loses spatial and geometric details, especially at boundaries. 
The BEM path preserves captures boundary details, but lacks global semantic information.
As the features of the two paths differ in representation level, simply averaging these features may cause information conflict and segmentation errors.  
To solve this problem, we propose the feature fusion module to solve information conflict, and achieve better segmentation accuracy.
As shown in Figure~\ref{fig:overview}(f), we concatenate features $F_{\text{PMT}}$ and $F_{\text{BEM}}$. A relationship matrix $\mathbb{R}^{C \times PN \times PN}$ is obtained by inputting it into the convolutional network ($C$ is the number of channels, $PN$ the number of patches). We then perform channel-wise multiplication on the patches and channels corresponding to this relationship matrix, selecting features. Finally, we concatenate features to obtain the final feature representations.

\subsection{Loss Functions} 
Instead of supervising only the final segmentation map, we jointly supervise all three parts: $L_{\text{Semantic}}$, $L_{\text{Boundary}}$, and $L_{\text{Final}}$, as each serves a specific purpose in our design. The total loss $L_{\text{Total}}$ is the weighted combination of these parts:
\begin{equation}
L_{\text{Total}} = \lambda_1 L_{\text{Semantic}} +  \lambda_2 L_{\text{Boundary}} + \lambda_3 L_{\text{Final}},
\end{equation}
$\lambda_1$, $\lambda_2$ and $\lambda_3$ are balancing hyper-parameters.

\textbf{Semantic Loss.} 
To enable the PMT branch to learn better semantic information, we adopt the boundary relaxation loss (RL)~\cite{zhu2019improving}, sampling only part of the pixels within objects for supervision. Since pixel numbers on different surfaces vary greatly, we also use focal loss (FL) for optimization. The semantic loss is:
\begin{equation}
L_{\text{Semantic}} =\alpha_1 L_{\text{FL}} + \beta_1 L_{\text{RL}}. 
\end{equation}

\textbf{Boundary Loss.} Boundary pixel prediction faces a class imbalance problem due to fewer boundary pixels compared to non-boundary pixels. Using the weighted cross-entropy loss alone often yields coarse segmentation results. 
To solve this problem, we employ both binary cross-entropy (BCE) and dice loss (DL)~\cite{milletari2016v} to optimize boundary learning. The boundary loss is:
\begin{equation}
L_{\text{Boundary}} =\alpha_2 L_{\text{DL}} + \beta_2 L_{\text{BCE}}.
\end{equation}

\textbf{Final Loss.} In UHR segmentation tasks, using only pixel-level cross-entropy supervision can deteriorate detailed structural information. To better capture objects with significant size differences, we use cross-entropy loss (CE) and focal loss (FL) to supervise final inference results. The final loss is:
\begin{equation}
L_{\text{Final}} = \alpha_3 L_{\text{FL}} + \beta_3 L_{\text{CE}} 
\end{equation}

\begin{table}[t]
  \centering
  \caption{Comparisons with the state-of-the-art methods on the DeepGlobe dataset. $\uparrow$ means higher is better, $\downarrow$ means lower is better. * means cropping the UHR image to small patches and predicting the results separately.}
  \resizebox{0.9\linewidth}{!}{
  \begin{tabular}{ll|cccc}
    \toprule
       & \multirow{2}{*}{\textbf{Method}}  & mIoU & F1 & Acc & Mem \\
       &   & (\%)$\uparrow$ & (\%)$\uparrow$ & (\%)$\uparrow$ & (M)$\downarrow$ \\
       \midrule
    \parbox{2.5mm}{\multirow{11}{*}{\rotatebox[origin=c]{90}{\textbf{Generic}}}}
    & U-Net* & 37.3  & -   & -  &  \textbf{949}  \\
    & DeepLabv3+* & 63.1 & -  & - & 1279    \\
    & FCN-8s* & 71.8  & 82.6 & 87.6 & 1963  \\
    & U-Net & 38.4  & -   & -  & 5507  \\
    & ICNet  & 40.2  & - & - & 2557    \\
    & PSPNet & 56.6 & - & - & 6289  \\
    & DeepLabv3+ & 63.5 & -  & - & 3199   \\
    & FCN-8s & 68.8  & 79.8 & 86.2 & 5227  \\
    & BiseNetV1 & 53.0 & - & - & 1801  \\
    & DANet & 53.8 & - & - & 6812  \\
    & STDC & 70.3 & - & - & 2580  \\
    \midrule
    \parbox{2.5mm}{\multirow{12}{*}{\rotatebox[origin=c]{90}{\textbf{UHR}}}}
    & CascadePSP & 68.5 & 79.7 & 85.6 & 3236   \\
    & PPN & 71.9 & - & - & 1193  \\
    & PointRend & 71.8 & - & - & 1593  \\
    & MagNet & 72.9 & - & - & 1559  \\
    & MagNet-Fast & 71.8 & - & - & 1559  \\
    & GLNet  & 71.6 & 83.2 & 88.0  & 1865 \\
    & ISDNet & 73.3 & 84.0 & 88.7  & 1948  \\
    & FCtL  & 73.5 & 83.8 & 88.3 & 3167  \\ 
    & WSDNet & 74.1 & 85.2 & 89.1 & 1876 \\
    & GeoAgent  & 75.4 & 85.3 & 89.6 & 3990   \\
    & GPWFormer  & 75.8 & 85.4 & 89.9 & 2380  \\
    \rowcolor{ourscolor}
    & BPT (Ours) & \textbf{76.6} & \textbf{85.7} & \textbf{90.1} & 2074  \\
  \bottomrule
\end{tabular}}
    \label{sota_deepglobe}
\end{table}

\begin{table}[tb]
    \centering
    \caption{Comparisons with the state-of-the-art methods on the Inria Aerial dataset.}
    \scalebox{0.9}{\begin{tabular}{l l c c c c}
      \toprule
      & \textbf{Method} & 
      \begin{tabular}[c]{@{}c@{}} mIoU \\ (\%)$\uparrow$ \end{tabular} & 
      \begin{tabular}[c]{@{}c@{}} F1 \\ (\%)$\uparrow$ \end{tabular} &  \begin{tabular}[c]{@{}c@{}} Acc \\ (\%)$\uparrow$ \end{tabular} & \begin{tabular}[c]{@{}c@{}} Mem \\ (M)$\downarrow$ \end{tabular}  
      \\ \midrule
    
    \parbox{2.5mm}{\multirow{3}{*}{\rotatebox[origin=c]{90}{\textbf{Generic}}}}
    & DeepLabv3+ & 55.9 & -  & - & 5122    \\
    & FCN-8s & 69.1  & 81.7 & 93.6 & \textbf{2447} \\
    & STDC & 72.4 & - & - & 7410  \\
    \midrule
    \parbox{2.5mm}{\multirow{8}{*}{\rotatebox[origin=c]{90}{\textbf{UHR}}}}
    & CascadePSP & 69.4 & 81.8 & 93.2 & 3236   \\
    & GLNet  & 71.2 & - & -  & 2663 \\
    & ISDNet & 74.2 & 84.9 & 95.6  & 4680  \\
    & FCtL & 73.7 & 84.1 & 94.6 & 4332  \\ 
    & WSDNet & 75.2 & 86.0 & 96.0 & 4379 \\
    & GeoAgent  & 76.0 & 86.1 & 96.4 & 5780  \\
    & GPWFormer  & 76.5 & 86.2 & 96.7 & 4710  \\
    \rowcolor{ourscolor}
    & BPT (Ours)  & \textbf{77.1} & \textbf{86.3} & \textbf{96.8} & 4450  \\
     
    \bottomrule
    \end{tabular}}
    \label{sota_inria}
\end{table}

\section{Experiments}

\subsection{Experimental Setup}

\subsubsection{Datasets and Evaluation Metrics.} 
To validate our method's effectiveness, we conducted experiments on five UHR image datasets: DeepGlobe~\cite{demir2018deepglobe}, Inria Aerial~\cite{maggiori2017can}, CityScapes~\cite{cordts2016cityscapes}, ISIC~\cite{tschandl2018ham10000}, and CRAG~\cite{graham2019mild}.

The DeepGlobe dataset comprises 803 UHR images, split into 455/207/142 for training, validation, and testing, respectively. Each image is 2448 $\times$ 2448 pixels, with annotations for seven landscape classes. 
The Inria Aerial dataset includes 180 UHR images (5000 $\times$ 5000 pixels) with binary masks for building/non-building areas, divided into 126/27/27 for training, validation, and testing. 
The Cityscapes dataset contains 5000 images with 19 semantic classes, split into 2979/500/1525 for training, validation, and testing. 
The ISIC dataset consists of 2596 UHR images, divided into 2077/260/259 for training, validation, and testing. 
The CRAG dataset comprises 213 images with glandular morphology annotations, split into 173 for training and 40 for testing, with an average size of 1512 $\times$ 1516.

In all experiments, we follow common practices~\cite{ji2023guided} by adopting mIoU, F1 score, and pixel accuracy (Acc) to evaluate segmentation performance, with mIoU being the primary metric. We also assess efficiency through GPU memory cost, measured using the "gpustat" tool with a mini-batch size of 1.

\subsubsection{Baselines.}
We compare BPT with several representative baselines. Some methods are designed for UHR images (denoted as "UHR") and others are not ("Generic"). Generic segmentation baselines include 
U-Net~\cite{ronneberger2015u}, 
PSPNet~\cite{zhao2017pyramid}, 
DeepLabv3+~\cite{chen2018encoder}, 
FCN-8s~\cite{long2015fully},
BiseNetV1~\cite{yu2018bisenet},
BiseNetV2~\cite{yu2021bisenet},
DANet~\cite{fu2019dual}, 
STDC~\cite{fan2021rethinking}, while UHR segmentation baselines include CascadePSP~\cite{cheng2020cascadepsp},  
PPN~\cite{wu2020patch}, PointRend~\cite{kirillov2020pointrend}, 
MagNet~\cite{huynh2021progressive}, MagNet-Fast~\cite{huynh2021progressive}, GLNet~\cite{chen2019collaborative}, 
ISDNet~\cite{guo2022isdnet},
FCtL~\cite{li2021contexts}, WSDNet~\cite{ji2023ultra},
GPWFormer~\cite{ji2023guided},
GeoAgent~\cite{liu2023seeing},
DenseCRF~\cite{krahenbuhl2011efficient}, 
DGF~\cite{wu2018fast}, SegFix~\cite{yuan2020segfix}.
The results of baseline models are referenced from~\citep{ji2023guided}.

\subsubsection{Implementation.}
For PMT, patch size is set to 32$\times$32. The number of Patch-Merging Transformer Blocks and Patch Recovering Blocks is four. 
We pre-train the model on the ImageNet-1K dataset using AdamW with a momentum of 0.9 and a weight decay of $5 \times 10^{-2}$. The initial learning rate is $1 \times 10^{-3}$, and the learning rate follows the cosine schedule. Models are pre-trained for 300 epochs.
For segmentation training, we train models on MMSegmentation codebase with GTX 3090 GPUs. 
We optimize models using AdamW with an initial learning rate of $1 \times 10^{-4}$, decayed using a polynomial schedule with a power of 0.9. Hyperparameters are set as follows: $\alpha_1=0.6$, $\beta_1=0.4$, $\alpha_2=0.3$, $\beta_2=0.7$, $\alpha_3=0.5$, $\beta_3=0.5$, $\lambda_1=0.3$, $\lambda_2=0.3$, $\lambda_3=0.4$. 
Following common practices~\cite{ji2023guided}, maximum training iterations are set to 40k, 80k, 160k, 80k and 80k for Inria Aerial, DeepGlobe, Cityscapes, ISIC and CRAG, respectively.

    

\begin{table}[tb]
    \centering
    \caption{Comparisons with the state-of-the-art methods on the Cityscapes datasets.}
    \scalebox{0.9}{\begin{tabular}{l c c c}
      \toprule
     & \textbf{Method} & mIoU (\%)$\uparrow$ & Mem (M)$\downarrow$ 
      \\
    \midrule
    \parbox{2.5mm}{\multirow{4}{*}{\rotatebox[origin=c]{90}{\textbf{Generic}}}}
    & BiseNetV1 & 74.4 & 2147    \\
    & BiseNetV2 & 75.8 & 1602    \\
    & PSPNet    & 74.9 & 1584  \\ 
    & DeepLabv3 & 76.7 & \textbf{1468}  \\
    \midrule
    \parbox{2.5mm}{\multirow{9}{*}{\rotatebox[origin=c]{90}{\textbf{UHR}}}}
    & DenseCRF & 62.9 & 1575  \\
    & DGF      & 63.3 & 1727  \\
    & SegFix   & 65.8 & 2033  \\
    & MagNet   & 67.6 & 2007  \\
    & MagNet-Fast & 66.9 & 2007  \\
    & ISDNet & 76.0 & 1510  \\
    & GeoAgent  &  77.8 & 2953  \\
    & GPWFormer  & 78.1 & 1897  \\
    \rowcolor{ourscolor}
    & BPT (Ours) & \textbf{78.5} & 1686  \\
    \bottomrule
    \end{tabular}}
    \label{sota_cityscapes}
    \vspace{-0.2cm}
\end{table}

\begin{table}[t]
\centering
\caption{Comparisons with the state-of-the-art methods on the CRAG and ISIC datasets.}
\scalebox{0.93}{\begin{tabular}{l|c|c}
\toprule
\multirow{2}{*}{Method}   & \multicolumn{1}{c|}{ISIC}      & \multicolumn{1}{c}{CRAG}  \\ 
            & mIoU (\%)$\uparrow$   & mIoU (\%)$\uparrow$   \\ \midrule
PSPNet   & 77.0    & 88.6  \\
DeepLabV3+    & 70.5   & 88.9  \\
DANet     & 51.4     & 82.3     \\
GLNet     & 75.2     & 85.9       \\
GeoAgent     & 80.2   & 89.4    \\ 
GPWFormer     & 80.7   & 89.9    \\ 
\rowcolor{ourscolor}
BPT(Ours)    & \textbf{81.6}   & \textbf{90.9}     \\ 
 \bottomrule
\end{tabular}}
\label{sota_isic_crag}
\vspace{-0.5cm}
\end{table}

\begin{table*}[tb]
\centering
\caption{Ablation studies on the DeepGlobe, Inria Aerial, and Cityscapes datasets. 
}
\resizebox{\textwidth}{!}{%
\begin{tabular}{c|cccc|cc|cc|cc}
\toprule
\multirow{2}{*}{ExpID} & \multicolumn{4}{c|}{Methods} &  \multicolumn{2}{c|}{DeepGlobe} & \multicolumn{2}{c|}{Inria Aerial} & \multicolumn{2}{c}{Cityscapes} \\
\cline{2-11}
 & PMB & PRB & BPT & FFM & mIoU (\%)$\uparrow$ & Mem (\%)$\downarrow$ & mIoU (\%)$\uparrow$ & Mem (\%)$\downarrow$ & mIoU (\%)$\uparrow$ & Mem (\%)$\downarrow$ \\
\midrule
\#1 & \cmark & \cmark & \cmark & \cmark & 76.6& 2074& 77.1& 4450& 78.5& 1686\\
\#2 & \xmark & \xmark & \cmark & \cmark & 73.9(-2.7)& 2153& 74.8(-2.3)& 4592& 76.9(-1.6)&1755\\
\#3 & \xmark & \cmark & \cmark & \cmark & 75.0(-1.6)& 2100 & 75.8(-1.3)& 4562& 77.7(-0.8)& 1743\\
\#4 & \cmark & \xmark & \cmark & \cmark & 75.5(-1.1)& 2132 & 73.8(-1.0) & 4496& 77.8(-0.8)& 1700\\
\#5 & \cmark & \cmark & \xmark & \xmark & 75.5(-1.1)& 2018 & 76.1(-1.0)& 4364& 77.9(-0.6)& 1622\\
\#6 & \cmark & \cmark & \cmark & \xmark & 76.1(-0.5)& 2035 & 76.7(-0.4)& 4396& 78.3(-0.2)&1642\\
\bottomrule
\end{tabular}%
}
\label{tab:ablation}
\vspace{-0.5cm}
\end{table*}

\subsection{Experimental Results}
\textbf{DeepGlobe.}
As shown in Table~\ref{sota_deepglobe}, we compare our proposed BPT with the aforementioned baseline methods on the DeepGlobe test dataset. 
The results demonstrate that BPT surpasses all other methods in terms of mIoU, F1, and accuracy. Notably, BPT significantly outperforms GeoAgent and GPWFormer in mIoU, without extra gpu memory cost.

\textbf{Inria Aerial.}
Table~\ref{sota_inria} presents the comparisons on the Inria Aerial test dataset. This dataset poses a greater challenge due to its high resolution, with each image containing 25 million pixels—approximately four times that of DeepGlobe and finer foreground regions. 
The results indicate that our BPT outperforms the baselines by substantial margins in mIoU, while maintaining comparable memory costs.

\textbf{Cityscapes.}
To further validate the generality of our method, we present results on the Cityscapes dataset in Table~\ref{sota_cityscapes}. BPT consistently outperforms all other methods in mIoU, while also demonstrating efficient memory usage.

\textbf{ISIC and CRAG.}
The ISIC dataset's image resolution is comparable to that of Inria Aerial, whereas CRAG features lower resolution images than the other datasets. Table~\ref{sota_isic_crag} illustrates the experimental results, where BPT consistently achieves excellent performance across both datasets.

\subsection{Ablation Study}

We present ablation studies in Table~\ref{tab:ablation} to evaluate the effectiveness of the proposed Patch-Merging Transformer (PMT) and Boundary-Enhanced Module (BEM) on the DeepGlobe, Inria Aerial, and Cityscapes datasets.

\textbf{Effect of Patch-Merging Transformer (PMT).} 
PMT comprises Patch Merging Blocks (PMB) and Patch Recovering Blocks (PRB). 
To assess the impact of PMB, we built a baseline transformer network by replacing PMB with convolutional layer to downsample, as shown in ExpID \#3, resulting in significant performance drops (-1.6\%, -1.3\%, -0.8\% mIoU compared to ExpID \#1). 
Similarly, replacing PRB with a deconvolutional head led to performance declines of -1.1\%, -1.0\%, and -0.8\% mIoU (ExpID \#4). 
Comparing ExpID \#1 and \#2 demonstrates the effectiveness of PMT.

\textbf{Effect of Boundary-Enhanced Module (BEM).} 
BEM significantly enhances model performance with minimal memory increase (ExpID \#1 and \#5), highlighting the importance of boundary information. 
The Feature Fusion Module (FFM) is also crucial for guiding information fusion, resulting in improved accuracy (ExpID \#1 vs. \#6).

\begin{figure}[t]
    \centering
    \includegraphics[width=0.99\linewidth]{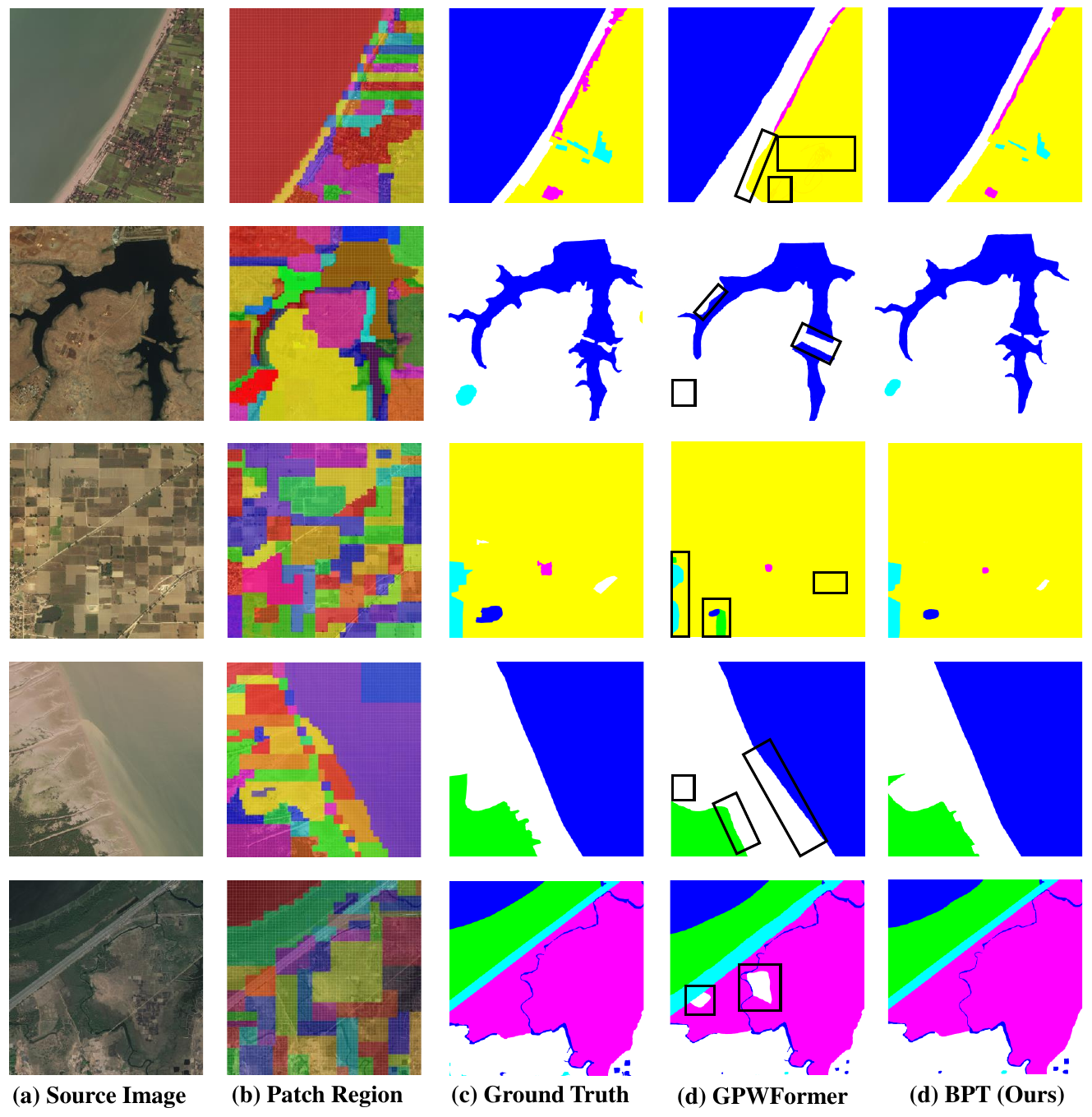}
    \caption{Qualitative analysis on the DeepGlobe dataset. 
    (a) Source image. (b) Patch tokens generated by PMT. (c) Ground-truth mask. (d) Results of GPWFormer. (e) Results of BPT (ours).}
    \label{fig:qualitative}   
    \vspace{-0.5cm}
\end{figure}

\subsubsection{Qualitative Analysis.}
To demonstrate the effectiveness of our proposed method, we conduct a qualitative analysis on the DeepGlobe dataset in Fig.~\ref{fig:qualitative}.
We visualize (a) the source image, (b) patch tokens generated by PMT, (c) the ground-truth mask, (d) results of the previous SOTA method, \ie GPWFormer, (e) results of  our proposed BPT.

\textbf{Analysis of Patch Merging.}
As shown in Fig.~\ref{fig:qualitative}(b), our method clusters patches of the same category into dynamic tokens of various shapes and sizes. Large areas of water, with minimal textual details, are represented by fewer tokens, while small, fine-grained areas use more tokens to preserve local information. Dynamic token allocation is crucial for preserving both global and local information without increasing computational cost.

\textbf{Comparisons with GPWFormer.} 
We compare our BPT with the previous state-of-the-art method, GPWFormer. Our BPT produces more precise results with better semantics and finer segmentation boundaries, showcasing its superior performance (Fig.~\ref{fig:qualitative}(d)(e)).

\section{Conclusion}
In this work, we propose a novel UHR image segmentation method, termed Boundary-Enhanced Patch-Merging Transformer (BPT). The Patch-Merging Transformer dynamically and adaptively merges patches to effectively capture both global semantic information and local fine-grained details. The Boundary-Enhanced Module (BEM) enhances segmentation accuracy by enriching fine details. Extensive experiments demonstrate that our BPT consistently outperforms existing methods on various UHR image segmentation benchmarks.

\section{Acknowledgments}
This work is supported by the Strategic Priority Research Program of Chinese Academy of Sciences (No.XDA28040500), the Natural Science Foundation of China (No.62302487), and two projects from the Science and Technology Innovation Program of Hunan Province (No.2024JJ9031 and No.2022RC4006).


\end{document}